\documentclass[preprint,12pt]{elsarticle}

\usepackage{graphicx}%
\usepackage{multirow}%
\usepackage{amsmath,amssymb,amsfonts}%
\usepackage{amsthm}%
\usepackage{mathrsfs}%
\usepackage[title]{appendix}%
\usepackage{xcolor}%
\usepackage{textcomp}%
\usepackage{manyfoot}%
\usepackage{booktabs}%
\usepackage{algorithm}%
\usepackage{algorithmicx}%
\usepackage{algpseudocode}%
\usepackage{listings}%
\usepackage{natbib}
\usepackage{amsmath}
\DeclareMathOperator*{\argmin}{arg\,min}

\journal{Expert Systems with Applications}
\biboptions{authoryear}

\begin{document}

\begin{frontmatter}


\title{Hierarchical Ensemble-Based Feature Selection for Time Series Forecasting}
\tnotetext[t1]{This work is in part supported by the Turkish Academy of Sciences Outstanding Researcher Program.}

\author[inst1]{Aysin Tumay\corref{cor1}}
\cortext[cor1]{Corresponding author.}
\ead{aysin.tumay@ug.bilkent.edu.tr}

\author[inst1]{Mustafa E. Aydin}
\ead{enesa@ee.bilkent.edu.tr}

\author[inst1]{Ali T. Koc}
\ead{atkoc@ee.bilkent.edu.tr}

\author[inst1]{Suleyman S. Kozat}
\ead{kozat@ee.bilkent.edu.tr}

\affiliation[inst1]{organization={Department of Electrical and Electronics Engineering},
            addressline={Bilkent University}, 
            city={Ankara},
            postcode={06800}, 
            country={Turkey}}

\begin{abstract}
We introduce a novel ensemble approach for feature selection based on hierarchical stacking for non-stationarity and/or a limited number of samples with a large number of features. Our approach exploits the co-dependency between features using a hierarchical structure. Initially, a machine learning model is trained using a subset of features, and then the output of the model is updated using other algorithms in a hierarchical manner with the remaining features to minimize the target loss. This hierarchical structure allows for flexible depth and feature selection. By exploiting feature co-dependency hierarchically, our proposed approach overcomes the limitations of traditional feature selection methods and feature importance scores. The effectiveness of the approach is demonstrated on synthetic and the well-known real-life datasets, providing significant scalable and stable performance improvements compared to the traditional methods and the state-of-the-art approaches. We also provide the source code of our approach to facilitate further research and replicability of our results.
\end{abstract}


\end{frontmatter}


\section{Introduction}
\subsection{Background}
\label{sec:backg}
We study feature selection for time series regression/prediction/forecasting tasks for settings where the number of features is large compared to the number of samples. This problem \citep{feat_sel_ts_1,feat_sel_ts_2,feat_sel_ts_3} is extensively studied in the machine learning literature as it relates to the infamous ``curse of dimensionality'' phenomenon, which suggests that machine learning models tend to struggle in cases where the number of samples is not sufficient given the number of features for effective learning from data \citep{BOLONCANEDO20191,some_other_curseofdim}. This results in over-training to obtain a model with high variance, i.e., low generalization ability \citep{friedman1997bias}.
This feature selection problem is even more prominent in non-stationary environments, e.g., for time series data or drifting statistics, where the trend, or the relationship between the features and the desired output changes significantly over time, making it challenging to identify relevant features.

Generally, the problem is addressed by i) considering all subsets of the features such as wrappers \citep{kohavi1997wrappers}, ii) using feature characteristics such as filters \citep{quinlan1986induction}, iii) embedding feature selection into the model learning process such as Lasso Linear Regression \citep{tibshirani1996regression}, Random Forest \citep{breiman2001random}, and iv) feature extraction methods such as Principal Component Analysis (PCA) \citep{pearson1901lines}. However, these methods are not effectively utilizing the information of the data, e.g., simply exploiting the ``dominant'' features without exploring others, and/or not scalable in reduction as the number of dimensions grows, i.e., computationally inefficient or not dynamic enough to adapt well as the domain of the features varies depending on the task. These widely used methods are also univariate, considering each feature once in calculating their importance. Directly evaluating all the subsets of the features for a given data becomes an NP-hard problem as the number of features grows, rendering such techniques, e.g., wrappers, computationally infeasible since the number of subsets reaches over a billion when the number of features exceeds thirty.
Embedding selection methodology into modeling, e.g., feature selection based on feature importance scores is also inadequate since, with limited data, these scores are unreliable giving vague explanations about gain or split-based selection for tree-based models \citep{natekin2013gradient}. One possible solution is to use ensemble or bagging techniques, where different machine learning algorithms are trained on different subsets of the feature vectors. However, this approach also leads to losing co-dependency information between features. Lastly, unsupervised feature extraction techniques such as PCA suffer from not incorporating valuable information from the underlying task where the original task, be it regression or classification, is supervised.

Here, we introduce a highly effective and versatile hierarchical stacking-based ensembling approach to this problem, where we first train an initial machine learning model using the dominant subset of the full features and then a second machine learning model in the hierarchy takes the remaining features as inputs using only the outcome of the first model on the cost function. Our method is promptly hierarchical since we train the initial model with the feature subset that is observed to be capturing the top importance scores in a baseline model. The rest of the features inputted to the models lower in the hierarchy are split based on domain expertise. The hierarchy increases until either the features exhaust and/or a user-controlled hierarchy depth is reached. Therefore, this generic structure allows for the depth of the hierarchy to be a design parameter, as well as the features used in each layer. By exploiting the co-dependency between features in a hierarchical manner, our approach addresses the limitations of traditional feature selection methods and provides more reliable results than feature importance scores. We build upon the substantial work demonstrated in the ensemble feature selection domain and illustrate the success of our approach on the synthetic and well-known real-world datasets in terms of accuracy, robustness, and scalability.

\subsection{Related Work}
Filters \citep{quinlan1986induction} and wrappers \citep{kohavi1997wrappers} are the prevalent traditional feature selection methods in use. While the former utilizes statistical tests such as chi-square, information gain, and mutual information, wrappers recursively do forward selection or backward elimination to the current feature subset according to an evaluation metric assigned to the model. The forward selection has a hard time finding good co-predictors while it is faster than backward elimination resulting in better scalability to larger datasets. In situations where wrappers overfit, filters are used with the knowledge of statistical tests \citep{das2001filters}. Filters are fast in computation, easy to be scaled to higher dimensional datasets, and independent of the model while the dependencies between features are ignored.  Wrappers, on the other hand, interact with the model and model the feature dependencies while they tend to require more computational resources since all feature subsets are tedious to try compared to filters. Saeys et al. \citep{saeys2007review} suggest more advanced methods such as an ensemble of feature selection methods and deep learning for feature selection. Moreover, Hancer \citep{Hancer2021ImprovedEvolutionary} proposes a wrapper-filter feature selection method using fuzzy mutual information that overcomes standard mutual information's limitations.   Our approach differentiates from wrappers, filters, and the methods of Saeys et al. and Hancer since we propose a multivariate solution that processes the groups of features by leveraging the codependency in the groups of features.

Boosting methods in feature selection as suggested by Das \citep{das2001filters} are equipped with boosted decision trees where the metric is information gain and weak learners are decision stumps. They perform well with the help of increasing the weight of each high-loss decision stump. This paper also proposes a hybrid method that uses a filter method for initial feature ranking and selection, followed by a wrapper method that evaluates the selected features using a classification model, combining the high accuracy of wrappers and time efficiency of filters \citep{saeys2007review}. Even though the hybrid model of wrappers and filters generates a more efficient model, our method is more time-efficient and utilizes the statistical importance of features as well as the prior knowledge of the side information. On top of these, the gradient boosting models are not generic for every loss function since these models require the hessian of the loss function to be nonzero. Our approach can integrate any external loss function in the middle of the system, independent of the boosting algorithm, which brings a novel solution to the problem.

The embedded methods, such as feature importance extracted by Random Forest \citep{breiman2001random} algorithms and unsupervised feature extraction methods, are not task-specific and might be inadequate to exploit domain knowledge in many areas. The feature selection method of RF determines the features that most reduce the impurity across all trees out of a random bag of features. The split strategy can be based on the Gini impurity of information gain, both yielding a univariate method for feature importance. 
While feature importance scores of tree-based models bring insight into the dataset, they ignore the correlations between features since each feature is individually analyzed.   

Overall, the contemporary literature addresses the feature selection either in a disjoint manner, i.e., in a model/domain-independent fashion, or via favoring the seemingly dominant features while failing to explore the majority of the features. Unlike previous studies, our model, for the first time in the literature, fully exploits the co-dependency between features via a hierarchical ensemble-based approach. This allows for an adaptive, i.e., dynamic feature selection, which is especially useful in nonstationary environments, e.g., in time series settings. The introduced architecture is generic, i.e., both the depth of the hierarchy and the base models employed are user-controlled depending on the nature of the high-dimensional dataset. As shown in our simulations, the introduced model provides significant improvements in performance as well as scalability over the well-known real life competition datasets compared to the traditional as well as the state-of-the-art feature selection approaches. We publicly share the implementation of our algorithm 
for both model design, comparisons, and experimental
reproducibility \footnote{https://github.com/aysintumay/hierarchical-feature-selection}.

\subsection{Contributions}
We list our main contributions as follows.

\begin{enumerate}
    \item We extend the limitations of traditional Gradient Boosting Machines where the loss functions with undefined hessian are flexibly injected as an external block of optimization in our proposed structure. 
    \item We overcome the issue of side information ignorance due to high dimensional data thanks to the hierarchical placing of domain knowledge acquired by domain expertise.
    \item Overcoming the curse of dimensionality, our hierarchical stacking approach is especially successful in small datasets with small instances and considerably high dimensions where the trend is hard to capture due to nonstationarity.
    \item We incorporate multivariate feature selection by maintaining the co-dependent feature pairs.
\end{enumerate}


\section{Material and methods}\label{sec:prelim}

\subsection{Problem Statement}\label{sec:prob_state}
All vectors in this paper are column vectors in lowercase boldface type. Matrices are denoted by uppercase boldface letters.  Specifically, $\boldsymbol{X} $ represents a matrix containing ${x}^{(k)}_{t}$, i.e., the  $k^{th}$ sequence of  vector $\boldsymbol{x}_{t}$, and ${x}^{(k)}$ in the $k^{th}$ column for each time $t$. $X_{i,j}$ denotes the element of $\boldsymbol{X}$ in the $i^{th}$ row and $j^{th}$ column.  Ordinary transposes of $\boldsymbol{x}_t$ and $\boldsymbol{x}$  are denoted as $\boldsymbol{x}^T_t$ and $\boldsymbol{x}^T$, respectively. The mean and standard deviation of ${x}_t^{(k)}$, i.e., the $k^{th}$ dimension of $\boldsymbol{x}_t$, are denoted by $\bar{{x}}_t^{(k)}$ and $\sigma({x}_t^{(k)})$, respectively. The covariance of a time series $\{y_t\}$ with its $k$ times delayed version is represented as $\gamma_y(t,t+k) = cov({y}_t, {y}_{t-k})$. The gamma function, having the property of $\Gamma(N) = (N-1)!$ generalizes the concept of a factorial to real and complex numbers.

We study feature selection in time series prediction of a sequence $\{y_t\}$. We observe this target sequence along with a side information sequence (or feature vectors) $\boldsymbol{x}_t$ each of which is of size $M$. At each time $t$, given the past information $\{y_k\}$, and $\boldsymbol{x}_k$ for $k \leq t$, we produce the output $\hat{y}_{t+1}$. Hence, in this setting, our goal is to find the relationship
\begin{equation*}
    \begin{aligned}
        \hat{y}_{t+1} = F_t\big(\{&y_{1}, y_{1},\ldots, y_t\}, \{\boldsymbol{x}_{1},\boldsymbol{x}_{2}\ldots, , \boldsymbol{x}_{t}\}\big),
    \end{aligned}
\end{equation*}
where $F_t$ is an unknown function of time, which models $\hat{y}_{t+1}$. We introduce a hierarchical nonlinear ensemble model for $F_t$ with an efficient feature selection procedure integrated in.
Throughout the training of the model, we suffer the cumulative loss
\begin{equation*}
    L = \sum_{t=1}^{N}\ell(y_{t}, \hat{y}_{t}),
\end{equation*}
where $N$ is the number of data points and $\ell$ can be, for example, the squared error loss, i.e., $\ell(y_{t}, \hat{y}_{t}) = (y_{t} - \hat{y}_{t})^2$.


Formally, at time step $t$, we have a matrix $\boldsymbol{X}$ with a length $N$ and dimensionality $M$, i.e., the $N$ data points of $\boldsymbol{X}$ form an Euclidean space of $M$-dimensions. Considering this hypersphere, we generalize the distance between data points in the $M$-dimensional hypersphere as $d \approx 2 \cdot r$, where $r$ is the radius of the hypersphere. If $M$ approaches $N$, the volume of the hypersphere, $V_n(r)$, given by:
\begin{equation}
\label{eq:volume}
\centering
 V_n(r) = \frac{\pi^{\frac{M}{2}}}{\Gamma\left(\frac{M}{2} + 1\right)} r^M
\end{equation}
increases exponentially, leading to a sparser data distribution characterized by a larger $d$ \citep{bellman1957dynamic}. 

We suffer data sparsity also due to non-stationarity, which is generally caused by trends and seasonality, while conducting time series forecasting for target sequence $\{y_t\}$.  
As such, trends and seasonality reduce the relative number of unique data points, which can lead to overfitting. Therefore, datasets with small sample sizes and dimensions larger than 30 are prone to overfitting. 

To this end, we propose a hierarchical ensemble-based feature selection method for the time series forecasting task to overcome overfitting and non-stationarity. We split $\boldsymbol{x}_t$ into $K$ non-intersecting feature subsets $\boldsymbol{s}_t^{(k)}, k = 1, ..., K$ based on domain knowledge or using certain heuristics such as feature importance metrics extracted from baseline model as demonstrated in our experiments. After this split, we train $K$ machine learning models, in a dependent manner, that take each  $\boldsymbol{s}_t^{(k)}$ as input in a hierarchical order by optimizing a cost function after each model until the $K^{th}$ one. The hierarchy lies in the order of $\boldsymbol{s}_t^{(k)}$s inputted to $K$ algorithms. We input the most dominant $\boldsymbol{s}_t^{(k)}$ based on feature importance scores of a baseline model to the $1^{st}$ model. Utilizing feature importance scores is harmless at this level since there is a significant leap between the side information and the selected feature subset. Then, the other features are split based on domain expertise and the context of the side information. 
Before describing our approach, we discuss the well-known approaches to this problem for completeness in the following.

\subsection{Common Approaches}\label{sec:problem_description}
There are many approaches for circumventing the curse of dimensionality, such as wrapper-based, embedded, filtering, and ensemble methods. We employ backward selection for wrapper-based, Pearson correlation coefficient for filtering, and LightGBM feature importance scores for embedded models. Let us denote a subset of the full feature vector $\boldsymbol{x}_t$ as \(\boldsymbol{s}^{(k)}_t\), \(k = 1, \ldots, 2^M\), i.e.,  given $M$ features, there exist $2^M$ different subsets of the feature set, which may include one or more features from \(\boldsymbol{x}_t\).

\subsubsection{Wrapper-based Methods}
As a greedy method, the well-established wrappers have an optimization objective over the validation loss that finds the best-performing feature subset \(\boldsymbol{s}^*_t\) as follows:

\begin{align}\label{eq:wrap}
    \boldsymbol{s}^*_t = \argmin_{\boldsymbol{s}^{(k)}_t \subset \boldsymbol{x}_t} L_{val} = \argmin_{\boldsymbol{s}^{(k)}_t \subset \boldsymbol{x}_t} \sum_{t' = t_1}^{t_2} \ell(y_{t'}, f_{\boldsymbol{w}}(y_{t'}, \boldsymbol{s}^{(k)}_{t'})).
\end{align}
subject to \(k = 1, \ldots, 2^M\), where $\{y_{t}\}_{\{t_1:t_2\}}$ and $\{s_{t}^{(k)}\}_{\{t_1:t_2\}}$ are the target sequence and the feature subset ${s_{t}^{(k)}}$ of the validation set between $t_1, t_2 \in \mathbb{Z}, \quad 1 <t_1 < t_2 < t$, \(f_w\) is a machine learning model trained on the \(k^{\text{th}}\) feature subset \(\boldsymbol{s}^{(k)}_t\) with parameters $\boldsymbol{w}$, and \(\ell\) is the loss function of the model. The algorithm can iterate through each feature subset \(\boldsymbol{s}^{(k)}_t\), seeking to maximize the performance of the machine learning model on the validation set. We employ a backward selection structure where (\ref{eq:wrap}) objective is conditioned to eliminating $\boldsymbol{x}_t$s one by one in each iteration. Although the objective is calculated on the validation set, the elimination of one feature in each iteration hurts the feature information in co-dependent feature pairs since the algorithm is unaware of the context of ${s_{t}^{(k)}}$s.
Naturally, due to computational complexity issues, wrappers are hardly used in a complete form in real-life applications in most cases.

\subsubsection{Embedded Methods}
As another approach in the literature, embedded methods perform feature selection as a part of the model construction process. Examples of these methods include Random Forests, and Gradient Boosting Trees.
Tree-based models eliminate features once or recursively (also known as Recursive Feature Elimination) based on their feature importance rankings and uses the remaining ones for training. One drawback of this method is that it is univariate, considering one feature at a time while calculating the scores. 

%

%
\begin{figure}
    \centering
    \includegraphics[width = 10cm, height = 5cm]{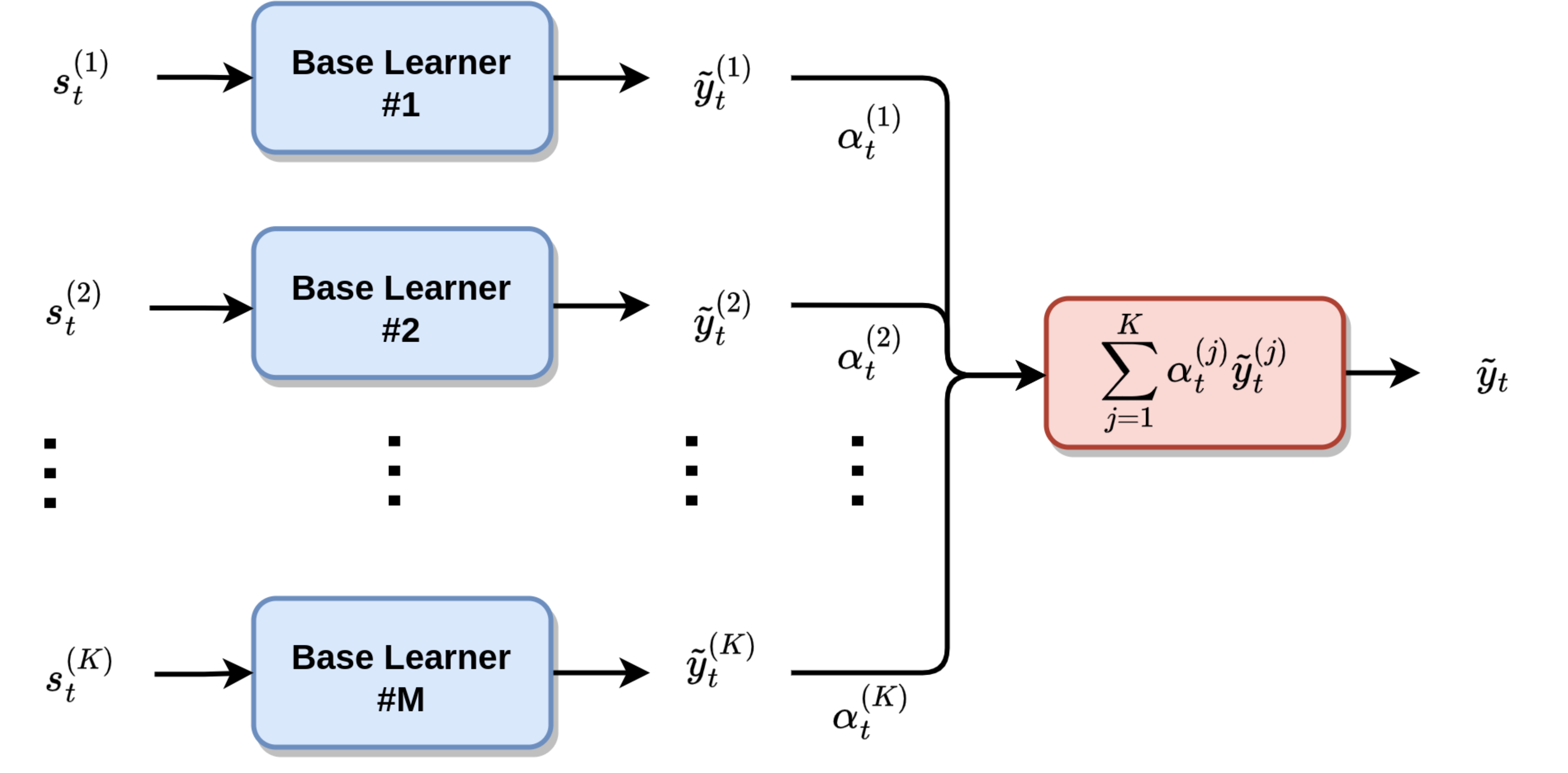}
    \caption{We have $K$ number of feature subsets used as inputs to $K$ base models (blue). Then, we combine base learners with $\boldsymbol{\alpha}_t^{(i)}$ for final prediction (pink). }
    \label{fig:ensemble}
\end{figure}

\subsubsection{Filtering Methods}

Another traditional method of feature selection is filtering. Unlike other methods, filtering relies on statistical measures instead of using machine learning algorithms. For instance, the score of the Pearson Correlation Coefficient \citep{pearson1896mathematical} between $x_{t^{'}}^{(k)}$ and $\{y_{t^{'}}\}$ for $t^{'} \leq t$ is $S(x^{(k)}) = {\text{cov}}(x^{(k)}, y) / (\sigma(x^{(k)}) \cdot \sigma(y))$.

 Filtering algorithm can incrementally form an optimal set $\boldsymbol{s}_t^*$ in terms of correlation with $m$ feature vectors from $\boldsymbol{x}_t$ based on maximum dependency. This is accomplished by discarding the lowest correlation score giving feature vectors from $\boldsymbol{x}_t$ in each iteration. 
 On the other hand, filtering methods do not inherently incorporate domain knowledge and cannot measure nonlinear dependency since they solely rely on statistical measures. Moreover, they are univariate, calculating the score of each feature one by one.

\subsubsection{Ensemble-based Methods}

 In this method, the predictions of machine learning models, also called base learners, are directly used to determine the weight vector that ensembles the base learners,  as shown in Figure \ref{fig:ensemble}. The version with two base learners is demonstrated in Algorithm \ref{alg:ens}.  All base learners take different feature subset vectors as input. Combining the predictions of $K$ base learners, denoted as ${{\tilde{y}}^{(i)}_t}, i = 1, \ldots, K$, the ensemble prediction is found as follows:

\begin{equation}
\centering
{{\tilde{y}}_t^E = \boldsymbol{\alpha}_t}^T{\boldsymbol{\tilde{y}}_t},
\end{equation}
where $\boldsymbol{\tilde{y}}_t = [{\tilde{y}}^{(1)}_t, ...,  {\tilde{y}}^{(K)}_t]^T$ is the base prediction vector of $K$ machine learning models, and $\boldsymbol{\alpha}_t$ is the ensembling coefficient vector $\boldsymbol{\alpha}_t \in \mathbb{R}^K$. With affine-constraint optimization, the loss of ensemble models is subject to $\boldsymbol{\alpha}_{\boldsymbol{s}_t^{(i)}}^T \boldsymbol{1}= 1$:

\begin{equation}
\centering
\operatorname*{min}_{{\alpha}_{\boldsymbol{s}_t^{(i)}} \in \mathbb{R}^K} \ell\big({y}_{t}, \sum_{i=1}^K {\alpha}_{\boldsymbol{s}_t^{(i)}}^{(i)} \,{\tilde{y}}_{t}^{(i)}\big).
\end{equation}
The optimization hyperspace is $(K-1)$-dimensional when the $K^{th}$ component of $\boldsymbol{\alpha}_t$ complements the sum of the entire vector to be 1. Therefore, the subject of the minimization changes into ${\alpha}_{\boldsymbol{s}_t^{(i)}} = 1 - \boldsymbol{\alpha}_{\boldsymbol{s}_t^{(i)}}^T\boldsymbol{1}$. In this sense, conventional ensemble methods can be computationally expensive, and more significantly, since they are independently trained, they cannot exploit the co-dependency between feature subsets. 

\begin{algorithm}
\caption{Ensemble Model}
\label{alg:ens}

\begin{algorithmic}[1]
    \State ${\tilde{y}}^{(1)}_{t} = g(\boldsymbol{s}_t^{(1)}, {y}_t)$ \Comment{Train base learner $g$ with $\boldsymbol{s}^{(1)}_t$.}
    \State ${\tilde{y}}^{(2)}_{t} = h(\boldsymbol{s}^{(2)}_t, {y}_t)$ \Comment{Train base learner $h$ with $\boldsymbol{s}^{(2)}_t$.}
    \For{$t = 1$ to $N$} \Comment{Iterate through each time step $t$.}
        
        \State $\text{min\_loss} \gets \infty $
        \State $\tilde{\alpha}_t \gets 0$
        \For{$\alpha = 0$ to $1$}
            \State ${\tilde{y}}_{t} = \alpha \cdot {\tilde{y}}^{(1)}_{t} + (1- \alpha) \cdot {\tilde{y}}^{(2)}_{t}$
            \State $\text{loss} = \ell({y}_{t}, \tilde{y}_{t})$
            \If {$\text{loss} < \text{min\_loss}$}
                \State $\text{min\_loss} = \text{loss}$
                \State $\tilde{\alpha}_t \gets \alpha$
            \EndIf       
        \EndFor 
    \EndFor
\end{algorithmic}
\end{algorithm}

\subsection{The proposed model}\label{sec:proposed}
\begin{figure*}[t]
    \centering
   \includegraphics[height = 3.3cm]{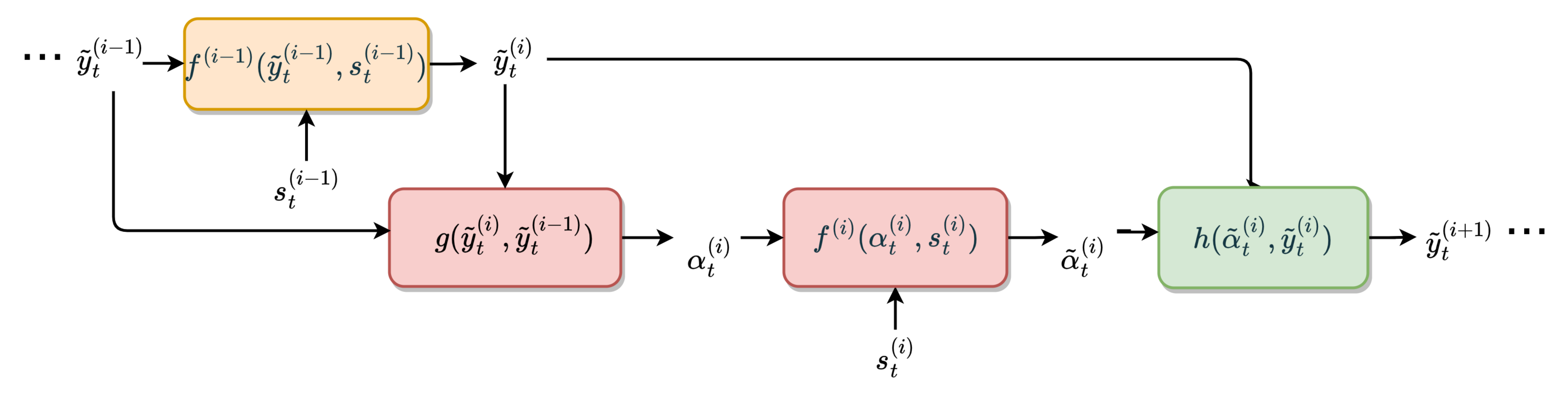}
    \caption{We have 2 feature subsets that are inputted to 2 different models in a hierarchical order. The first layer (orange) takes the $y_t$-related features as input. In the next step, ${\alpha}_t^{(i)}$ is generated with cost optimization. Then, the second layer (pink) predicts ${\alpha}_t^{(i)}$. Finally, the second layer predictions (green) are generated by combining ${\Tilde{\alpha}}_t^{(i)}$ and ${\Tilde{y}}^{(i)}_{t} $.}
    \label{fig:hierarchical_model}
\end{figure*}

To overcome the limitations of traditional feature selection methods, we propose a hierarchical ensemble-based approach involving $K$ distinct machine learning models organized into $K$ levels.  Figure \ref{fig:hierarchical_model} illustrates two sample successive layers of the structure. Each machine learning model takes the output of the previous layer ($f^{(i-1)}$ in Figure \ref{fig:hierarchical_model}). Then, the latter model ($f^{(i)}$ in Figure \ref{fig:hierarchical_model}) generates the predictions of the optimized weights that scale the last output. Each model operates on a different subset of $\boldsymbol{x}_t$, which may include exogenous information, also called the side-information, and features derived from the past information of the target sequence $\{y_t\}$, $\{y_{t-j}\}, j = 1, \ldots, N-1$, denoted as $\boldsymbol{s}_t({y})$. 

While the hierarchical ordering of the models is guided by domain expertise, we propose that the first level should exclusively comprise $\{{y}_{t-j}\}$ (and features derived using them, e.g., their rolling means) referred to as ${y}_t$-related features as shown in different time series prediction papers \citep{Arik_1,Arik_2}. The reason is that these past target values exhibit higher importance scores than the side information sequences, thereby dominantly influencing predictions. In fact, many machine learning-based time-series models suffer from ``overfitting'' to the ${y}_t$-related features and ignore most of the features \citep{ml_models_favoring_yt_relateds}. Subsequent levels can incorporate the side information sequences $\boldsymbol{s}_t^{(k)} \subset \{\boldsymbol{x}_t\} \setminus \boldsymbol{s}_t({y})$ for $k=1, \ldots, K-1$. We next describe the layers in the hierarchy and the optimization procedure thereof. 

\subsection{Description of Layers}

To explicate the layers composing the introduced architecture, in Figure \ref{fig:hierarchical_model}, we present a snapshot of the generic model with all the major components in action; these represent the core operations in the overall hierarchy which are repeated in succession. In Figure \ref{fig:hierarchical_model}, we observe four main components in transitioning from layer $i-1$ to $i$: two machine learning models (left and middle right), a cost optimization function (middle left), and a linear superposition function (right). The leftmost machine learning model, $f^{(i-1)}$, is fed with $(i-1)\textsuperscript{th}$ restricted side information sequence $\boldsymbol{s}_t^{(i-1)}$, which corresponds to the ``feature bagging'' technique for addressing the bias-variance trade-off \citep{breiman2001random}, as well as with the \emph{refined} predictions of the previous layer, which we will elaborate on in the following. \textbf{There is {no restriction} on what this learning model could be, nor on the loss function it aims to minimize.}

After the learning is finished, the predictions of the model are acquired and passed onto the cost optimization block (middle left part in Figure \ref{fig:hierarchical_model}). Therein lies the novelty of our algorithm, as we, unlike the usual boosting procedure, e.g., LightGBM uses, do not transmit these predictions as is to the next model in the chain but instead subject it to a weighting. To minimize the \emph{final} loss, the cost optimization function $g$ generates a weight sequence $\alpha_t^{(i)}$ that scales the previous prediction sequence, $\tilde{y}_t^{(i-1)}$. Weighting the instances is necessary for our setup since $\boldsymbol{s}_t^{(i-1)}, i = 2,\ldots, K+1$ are designed to inhibit information about the direction of the growth of predictions, i.e., binary energy load take and load give instructions that guide the degree of undershooting or overshooting in the next prediction. Therefore, surpassing the extra increase and decrease on the next time step predictions is prominent. Later, we show in our simulations that feature uninformative of the direction of the growth of predictions, i.e., date features, can also be merged as side information.

Another contribution of this method lies in the compatibility of any loss function in the cost optimization stage, which extends on the limitations of loss functions with non-practical second derivatives, e.g., the L1 loss.  The primary goal of the setup is extending GBMs with loss functions of non-practical second derivatives. A trivial and generic representation of this problem is L1 loss which we indeed employ in our cost optimization step. Extending to any loss function upon L1 loss is convenient with this setup. The details of the optimization of $\alpha_t^{(i)}$s are given in Section \ref{sec:Cost}. 

For context-awareness, we further feed these optimized $\alpha_t^{(i)}$s into another learning model, $f^{(i)}$, which uses the current context $\boldsymbol{s}_t^{(i)}$ for its training; we acquire the context-aware weights $\tilde \alpha_t^{(i)}$ out of it. Then, in the last superposition stage, a linear function $h$ scales the prediction of the leftmost model $f^{(i-1)}$ with $\tilde \alpha_t^{(i)}$s to obtain the \emph{refined} predictions, which are then fed to the next block in the series. The chain of blocks continues this way until $i$ hits the user-defined hierarchy size parameter $K$. \textbf{Parallel to the choice of $f^{(1)}$, there are {no restrictions} on what the learning models could be, nor on the loss function it aims to minimize.} We demonstrate the stacking of 2 models, which is indeed enough for simulating the hierarchy since the context of $\boldsymbol{s}_t^{(i)}$s might not be different from each other in real cases. On top of that, only separating feature subset of $\{y_{t-j}\}$ and plugging it on top of the hierarchy prevents $f^{(i)}, i = 2, \ldots, K+1$ to diverge too much from $\tilde{y}_t^{(1)}$. The hierarchical flow of the model is depicted in Algorithm \ref{alg:cap}. 

\begin{algorithm}
\caption{Hierarchical Ensemble-Based Method}
\label{alg:cap}
\begin{algorithmic}[1]
    
    \State $\boldsymbol{s}^{(i-1)}_t \gets \boldsymbol{s}_t({y}) \in \boldsymbol{X}, \text{ where } t= 1,2, \dots, N, i= 2, 3, \dots ,K+1$
    \State ${\tilde{y}}^{(i)}_t \gets f^{(i-1)}(\boldsymbol{s}^{(i-1)}_t,\tilde{y}^{(i-1)}$)
    \Procedure{Cost Optimization}{}
    \For{$t = 1, 2, \dots, N$}
        
        \State $min\_loss \gets \infty$
        \State $range = [1-\beta,1 + \beta], \text{ where } \beta \in [0, 1] \subset \mathbb{R}$

        \For{$\alpha \gets 1-\beta$ to $1 + \beta$}
            \State $\hat{y}^{(i)}_t = \alpha \tilde{y}^{(i)}_t $
            \State $loss = \ell(y_t, \hat{y}^{(i)}_t)$
            \If{$loss < min\_loss$}
                \State $min\_loss = loss$
                \State $\alpha_t^{(i)} = \alpha$
            \EndIf       
        \EndFor 
    \EndFor
    \EndProcedure
    \State $\boldsymbol{s}^{(i)}_t  \gets \boldsymbol{s}^{(i)}_t \subset  \{\boldsymbol{x}_t\} \setminus  \boldsymbol{s}_t({y})$
    \State ${\tilde{\alpha}}_t^{(i)} \gets f^{(i)}(\boldsymbol{s}^{(i)}_t, {\alpha}_t^{(i)})$
    \State ${\Tilde{y}}^{(i+1)}_t \gets {\tilde{\alpha}}_t^{(i)}  {\Tilde{y}}^{(i)}$
    
\end{algorithmic}
\end{algorithm}

Formally, in the transition from $(i-1)\textsuperscript{th}$ to $i\textsuperscript{th}$ layer of the algorithm, we first generate $(i-1)\textsuperscript{th}$ model's predictions for each time $t$, denoted as ${\tilde{y}}^{(i-1)}_{t}$, where $i = 2, \ldots, K$, inputting $\boldsymbol{s}_t^{(i-1)}$ into base learner $f^{(i-1)}$. Subsequently, we deduce the weight sequence ${\alpha}^{(i)}_{t}$ to refine ${\tilde{y}}^{(i - 1)}_t$ by optimizing iteratively in a loop as depicted in Algorithm \ref{alg:cap} for:
\begin{equation}
\centering
{\hat{y}}^{(i)}_t = {{\alpha}_t^{(i)}} {\tilde{y}}_t^{(i)},
\end{equation}
where $\alpha_t^{(i)} \in [0, 2]$ for each time $t$. 
As depicted in the middle left pink block in Figure \ref{fig:hierarchical_model}, we determine ${\alpha}^{(i)}_{t}$ with an optimization loop to minimize the loss at each time $t$ in layer $i$ as,
\begin{align}\label{eq:cost}
    L_t = \ell({y}_{t}, {\hat{y}}_{t}^{(i)}) &= \ell({y}_{t}, {{\alpha}_t^{(i)}} {\tilde{y}}_{t}^{(i)}).
\end{align}
In general, $\ell$ need not be differentiable. Unlike ensemble methods, we optimize $\ell$ with ${\Tilde{y}}^{(i)}_{t}$ and ${\Tilde{y}}^{(i-1)}_{t}$ by scaling ${\Tilde{y}}^{(i-1)}_{t}$ at each time $t$ with ${\alpha}^{(i)}_t $.  The cost optimization step is further elaborated in Section \ref{sec:Cost}. In this sense, in the $i^{th}$ layer, we update the output of the $(i-1)^{th}$ layer using the features that belong to the $i^{th}$ layer. Therefore, every subset of features contributes to minimizing the final error similar to boosting or stacking. 

As shown in the middle right block in Figure \ref{fig:hierarchical_model}, we train the consequent model ($f^{(i)}$ in Figure \ref{fig:hierarchical_model}) with ${\alpha}^{(i)}_{t} $ and $\boldsymbol{s}_t^{(i)}$ to predict the ultimate weights ${{\tilde{\alpha}}}^{(i)}_{t}$, which are context-aware. In the green block at the end, we modify ${\tilde{y}}^{(i)}_t$ using ${\tilde{\alpha}}^{(i)}_{t}$ to incorporate the side information, leveraging the learned patterns from the error of the previous layer with a linear superposition function $h$, as represented by
\begin{equation}\label{eq:last_eq}
{\Tilde{y}}^{(i+1)}_{t} =  h(\tilde{y}_t^{(i)},   \tilde{\alpha}_t^{(i)}) ={{\tilde{\alpha}}^{(i)}_t} {\Tilde{y}}^{(i)}_t.
\end{equation}

In the following section, the cost optimization step between layers 1 and 2 in Algorithm \ref{alg:cap} is elaborated.

\subsection{Cost Optimization}
\label{sec:Cost}
The cost optimization refers to the iterative approach, in which we employ $\alpha_t^{(i)}$ from a determined range to modify $\tilde{y}_t^{(i)}$ in any layer $i$. We aim to find $\alpha_t^{(i)} \in \mathbb{R}$ for each time $t$. Finally, we have the following optimization objective:
\begin{equation}
\centering
\operatorname*{min}_{{\alpha}_{t} \in \mathbb{R}} \ell\big({y}_{t}, {\alpha}_{t}^{(i)} \,{\tilde{y}}_{t}^{(i)}\big),
\end{equation}
where $ \ell$ can be any loss subject to $ {\alpha}_t^{(i)} \in [0, 2]$ for each time $t$ such that if a weight is in the range $[0, 1]$ the prediction is effectively downscaled, and similarly for the range $[1, 2]$, we employ upscaling to leverage robustness. This procedure aims to be flexible so that the algorithm can be scaled to any domain-specific loss function.  The ``Cost Optimization" section in Algorithm \ref{alg:cap} and the leftmost pink block in Figure \ref{fig:hierarchical_model} shows the structure of the cost function.

To understand the contribution of our cost optimization approach, we compare it with the loss structure of a powerful tree-based model, LightGBM \citep{ke2017lightgbm}, where the second derivatives are effectively employed. Using the well-known Random Forest algorithm would not be fair to compare the proposed approach as the base model in each hierarchical step since one of our objectives is to solve the non-practical second derivative problem of GBMs. 
LightGBM algorithm is adjusted to generate the splits based on decision trees. Moreover, LightGBM's effective feature bundling property claims a solution for uninformative feature pairs in feature selection tasks. During the training process, LightGBM iteratively updates the model by minimizing the
chosen cost function.  This process is typically performed using
gradient-boosting techniques. Choosing the base model of LightGBM as a decision tree, the leaf split finding operation is completed with the high-level insight provided by the hessian and gradient of the loss function. Defining the gradient as 
\begin{equation}
{g}_{t}(x) = \mathit{E}_y \left[ \frac{\partial \ell(y_t, f(x_t, \theta, y_t)) }{\partial f(x_t, \theta, y_t)} \,|\, x \right]_{f(x_t, \theta, y_t) = \hat{f}_{t-1}(x_t, \theta, y_t)},
\label{eq:gradient}
\end{equation}
and the hessian is defined as
\begin{equation}
{h}_{t}(x) = \mathit{E}_y \left[ \frac{\partial^2 \ell(y_t, f(x_t, \theta, y_t)) }{\partial f(x_t, \theta, y_t)^2} \,|\, x \right]_{f(x_t, \theta, y_t) = \hat{f}_{t-1}(x_t, \theta, y_t)}.
\label{eq:hessian}
\end{equation}
If we use the customized loss function of LightGBM, it would calculate the gradient and hessian of \( \ell \) with respect to \( \tilde{y}_t^{(i)} \) to determine the direction and magnitude of the updates. As an example, the negative gradient for L1 loss is given by 
\begin{equation}
g_t^{(i)} = \frac{\partial \ell(y_t, \alpha_t \tilde{y}_t^{(i)})}{\partial \tilde{y}_t^{(i)}} = \alpha_t,
\label{eq:negative_gradient}
\end{equation}
which leads to the hessian being impractically 0, in a form that LightGBM cannot natively process. Therefore, the hessian of the loss function should be nonzero while working with a custom function. 

This user-friendly cost optimization step has controlled parameters as the \textbf{number of iterations, and range of the weights}.
In our application, we bring another approach to make it convenient to embed any custom loss into our base learners, e.g., LightGBM. In our case, we define the upper and lower limit of $\alpha_t^{(i)}$ range value by determining $\beta \in \mathbb{R}$, which is also a tuned parameter in the range of $[0,1]$, for $1+\beta$ to
be the higher limit and $1-\beta$ be the lower limit of the chosen $\alpha_t^{(i)}$ value. The number of iterations of the loop is determined based on the step size of the range $[1-\beta,1+\beta]$, which is another tuned parameter. As depicted in Algorithm \ref{alg:cap}, we find the optimal $\alpha_t^{(i)}$ in a greedy process aiming to minimize any custom loss function directly for each time step $t$. Therefore, the objective in (\ref{eq:cost}) is employed. 
After this step, $\tilde \alpha_t^{(i)}$ vector is inputted to the following model depicted as $f^{(i)}$ in Figure \ref{fig:hierarchical_model}.

By iterating the process in Algorithm \ref{alg:cap}, we can effectively search for the best combination of features and capture the correlation between them. For the sake of improving traditional feature selection methods, the iterative process allows us to optimize $K$ models simultaneously, as the predicted output from one model is used to enhance the other model.

In the next section, we illustrate the performance of our hierarchical ensemble-based model on synthetic and widely known real-life datasets.

\section{Experiments}\label{sec:simulations}
In this section, we illustrate the performance of our hierarchical ensemble-based model with 2 layers,  i.e., $K=2$, in comparison with other models on various well-known time series datasets. Initially, we introduce the models used for comparison. Then, we provide the performance of our model.

\subsection{Compared Models}
\label{sec:comp_method}
The simulations include five models that are labeled as \textit{Wrapper}, \textit{Ensemble}, \textit{Hierarchical Ensemble}, \textit{Embedded}, \textit{Baseline LightGBM}. The base model of the compared methods, as well as our proposed approach, is LighGBM since we aim to simulate the problem of undefined second derivatives of objective functions, which require the use of a GBM. On the other hand, we intend to achieve low mean square errors for each compared method. Therefore, we shift our choice of base model from Random Forest to LightGBM thanks to its brilliant aspect of exclusive feature bundling.

The first compared method, \textit{Wrapper}, which is described in Section \ref{sec:problem_description}, discards the feature that gives the least contribution to the model based on the L2 loss in each iteration. We omit one feature at each iteration of this algorithm. The \textit{Embedded} model described in Section \ref{sec:problem_description} only uses $\{y_{t-j}\}$  and features derived from $\{y_{t-j}\}$, e.g. rolling features, namely as $\boldsymbol{s}_t(y)$. The reason that $y_t$-related features are employed in the simulation as input to another model is to investigate if the most important features are enough without requiring domain knowledge. 
The \textit{Ensemble} model works based on Algorithm \ref{alg:ens} with 2 baseline models, mixing  $\boldsymbol{s}_t(y)$ and  $\{ \boldsymbol{x}_t\} \setminus \{  \boldsymbol{s}_t(y) \}$ with ${\alpha}_t^{(i)}$ which is chosen in an iterative process minimizing the L1 objective.  
Lastly, \textit{Baseline LightGBM} model refers to the model that uses the whole $\boldsymbol{x}_{t}$. With this model, we seek to find if feature selection for the datasets is necessary at all. Moreover, we expect to observe overfitting due to the high dimensionality and non-stationarity.   

We next introduce the evaluation strategy and elaborate on parameter tuning. 

\subsection{Evaluation Strategy and Parameter Tuning}
\label{sec:hyperparameter}
The evaluation metric used in the experiments is the mean square error. All experiments are iterated 200 times to ensure the reliability of the results. The synthetic dataset is generated 200 times with a random Gaussian noise. For the real-life dataset, we randomly sampled 200 out of 414 series of the hourly M4 Forecasting, the widely publicized competition dataset  \citep{MAKRIDAKIS2018802}. We obtained the cumulative sum of error between $\{\tilde{y}_t\}$ and $\{y_t\}$ for the $j^{th}$ experiment as follows:
\begin{equation}
\centering
{\mathit{l}}_t{^{(j)}} = ({y}^{(j)}_t - {\tilde{y}}^{(j)}_t)^2/N.
\end{equation}
Then, the average over the 200 trials is taken so as to eliminate the bias due to using a particular sequence as,
\begin{equation}
\centering
{\Bar{l}}_t = \frac{\sum_{j=1}^{200}{l}^{(j)}_t}{200}.
\end{equation}
Finally, the cumulative sum over time is taken to smoothen the results as follows:
\begin{equation}
\centering
{\Bar{l}}^{(ave)} = \frac{\sum_{t^{'} = 1}^{t}{\Bar{l}}_{t^{'}}}{t}.
\end{equation} 

To test the statistical significance of our proposed forecasts, we conduct a paired T-test between the MSE values of the compared methods and our proposed method. To setup the hypothesis,   $\mu_c$ and $\mu_p$ represent the mean of the compared and proposed methods, respectively. We state the hypothesis as $H_0: \mu_c \leq \mu_p, H_a: \mu_c > \mu_p $.

As another performance assessment strategy, we measure the average running times across 200 iterations of \textit{Wrapper} and \textit{Hierarchical Ensemble} only. The reason for not including the other methods is that they are not iterative since the feature selection is done beforehand without computation, i.e., \textit{Embedded} uses the hand-selected features with the highest feature importance score and \textit{Ensemble} merges \textit{Embedded} and the model with the rest of the features, thereby resulting in a much shorter running time. The non-iterative nature of these methods might lead to misinterpretation of the time efficiency of the proposed method.

About the experiment settings, the $\beta$ hyperparameter in the cost optimization step depicted in Algorithm \ref{alg:cap} is chosen as 0.33 for both experiments based on hyperparameter tuning with several trials as depicted in Table \ref{tab:hyperparameters}. Therefore, the range of $\alpha_t^{(i)}$ is $[0.66, 1.33]$. The number of iterations, which is numbers generated with the step size of the range, is also a tuned parameter and determined from the range depicted in Table \ref{tab:hyperparameters} after several trials.

\begin{table}[b]
    \centering
    \caption{Hyperparameter Space of Cost Optimization}
    \begin{tabular}{ccc} 
        \toprule
        &\textbf{Hyperparameter Space} & \textbf{Selected Value}\\
        \midrule
        $\beta$  & \{0.1, 0.2, 0.33, 0.5, 0.6, 0.9\} & 0.33\\
        Iterations & \{20,30, 40, 50, 60\} & 30 \\
        \bottomrule
    \end{tabular}
    \label{tab:hyperparameters}
\end{table}

In the next sections, we give analysis and experiments of synthetic and real-life datasets. As we simulate with $K=2$ models, we denote $\boldsymbol{s}_t^{(i-1)}$ as $\boldsymbol{s}_t^{(1)}$; likewise $\alpha_t^{(i)}$ as $\alpha_t$. 

\subsection{Experiments with Synthetic Dataset}
\begin{figure}[!htb]
    \centering
    \includegraphics[width=\columnwidth]{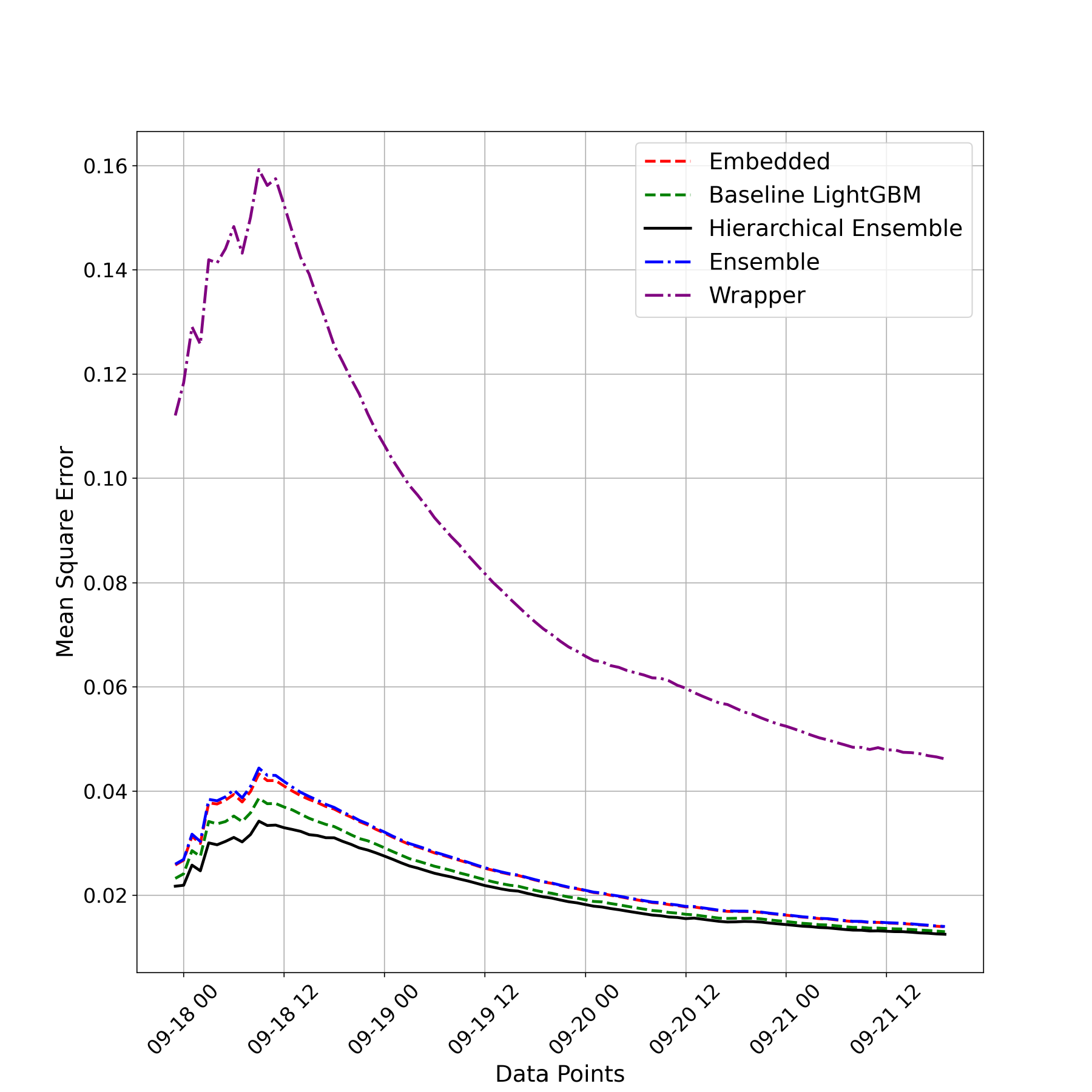}

    \caption{Comparison of the mean square error performances of \emph{Hierarchical Ensemble} (black), \emph{Ensemble} (blue), \emph{Base LightGBM} (green), \emph{Embedded} (red), \emph{Wrapper} (purple) for the synthetic dataset.}
    \label{fig:synthetic_simulation}
\end{figure}
The data is generated with an autoregressive moving average (ARMA) process of order $(4, 5)$, i.e.,
\begin{align}
\label{eq:arma}
\mathit{y_{t}} &= \sum_{i=1}^{4} \Phi_i \mathit{y_{(t-i)}} + \sum_{j=1}^{5} \theta_{j} \epsilon_{(t-j)} + \epsilon_{t},
\end{align}
where the autoregressive part is represented by the lagged values up to 4 times, and the moving
average is represented by the lagged error terms up to 5 times. The $\Phi$ and $\theta$ variables control the
strength of the lags \citep{box1970time}. In our setting, $\Phi = [0.4, 0.3, 0.2, 0.1]$, and $\theta = [0.65, 0.35, 0.3, -0.15,-0.3 ]$. We set up weights for five lags so the correlation is enough to prompt a seasonality and trend in the data. The Augmented Dickey-Fuller \citep{dickey1979distribution} test reveals the p-value as 0.2116, showing non-stationarity. Then, the series is transformed with a min-max scaler. We generate 500 samples of synthetic data to simulate the difficulty of base models to capture trends in time series data. As a result, \textbf{the synthetic data is equipped with non-stationarity, trend, and seasonality.}

We first generated the domain knowledge-representing feature subset $\{ \boldsymbol{x}_t\} \setminus \{  \boldsymbol{s}_t(y) \}$. For that, a binary classification feature set is designed with fully informative 26 features. The synthetically generated binary sequence $y_t^b$ has a class imbalance of $0.65$. The $y_{t}$ values obtained from \text{(}\ref{eq:arma}\text{)} that have indexes in the corresponding label 1 in $y_t^b$ are multiplied by $1.33$ to upscale, while others are multiplied by $0.66$ to downscale. This scaling is based on the experiment settings mentioned in the previous section. Therefore, we guarantee dependence on the generated domain knowledge series, which can provide a substantial amount of pattern. As the final step of generation, we added a Gaussian normal noise of $\mathit{N}(0, 0. 5)$.  To simulate the curse of dimensionality problem, the number of total features is set to 36, as demonstrated in Table \ref{table:data_info}, and most of the features are informative. The average p-value is also higher than $0.05$, which validates that our experiment settings are suitable for non-stationarity with a confidence level of 95\%. \textbf{The high dimensionality of the data is validated by the number of features being larger than 30.}

Upon generating the synthetic dataset with high dimensionality and non-stationarity, our dataset is well-suited to the problem statement in Section \ref{sec:prob_state}. 
Based on the performance plot in Figure \ref{fig:synthetic_simulation}, we highlight that our proposed method outperforms other compared models while the overall loss trend is decreasing. We confidently prompt that the $y_t$-related features, given to $\boldsymbol{s}_t^{(1)}$, find most of the patterns in the label while $\boldsymbol{s}_t^{(2)}$ finetunes the short-term patterns, as desired. Moreover, the descending trend indicates the robustness of our method.
The \textit{Wrapper} method is significantly the worst-performing model based on the MSE scores since it could not converge in a global optimum, using a suboptimal feature subset. From Table \ref{table:data_duration}, the extreme time consumption of this method compared to \textit{Hierarchical Ensemble} also verifies the time efficiency of our model. Since we generated highly co-dependent features, omitting one feature at a time reduces the informativeness of the features left. Hence, \textit{Wrapper} is the least efficient model among other compared methods. 
Additionally, \textit{Embedded} and \textit{Ensemble} models are significantly close in loss. One can say that \textit{Embedded} model generates long-term patterns successfully with $y_t$-related features. Therefore,  \textit{Ensemble} chooses $y_t$-related compared to date-related features. Even though the optimal feature subset is found by \textit{Embedded}, it is outperformed by \textit{Hierarchical Ensemble} since our method fully incorporates the information provided by codependent feature pairs rather than deciding whether to choose $y_t$-related or domain knowledge features.
Moreover, \textit{Baseline LightGBM} performs the closest to the proposed approach, which indicates the level of informativeness of the features. However, our proposed approach outperforms  \textit{Baseline LightGBM} since we overcome overfitting by training multiple models with less number of features while the other is overfitted easily due to a large number of features.

\begin{table}[t]
    \caption{Statistics of the Datasets}
    \label{table:data_info}
        \begin{tabular}{cccc}
            \hline
            \textbf{Dataset} & \textbf{Sample Size} & \textbf{Feature Size} & \textbf{Average p-value} \\
                        \hline

            Synthetic & 500 & 36 & 0.2376 \\

            M4 Hourly Forecasting & 1001 & 33 & 0.36856 \\
            \hline
        \end{tabular}
\end{table}

\begin{table}[b]
    \centering
    \caption{Paired t-test results for Mean Squared Error (MSE) - Synthetic Data}
    \begin{tabular}{lcc} 
        \toprule
        Method & T-statistic & P-value \\
        \midrule
        Baseline LightGBM & 12.264 & $8.28 \times 10^{-22}$ \\
        Ensemble & 16.667 & $1.16 \times 10^{-30}$ \\
        Embedded & 17.512 & $3.05 \times 10^{-32}$ \\
        Wrapper & 22.031 & $4.81 \times 10^{-40}$ \\
        \bottomrule
    \end{tabular}
    \label{tab:hypothesis_syn}
\end{table}

To test the statistical significance as introduced in Section \ref{sec:hyperparameter}, we input the first samples of MSE as the compared methods and the second as the proposed method. Table \ref{tab:hypothesis_syn} demonstrates that we have enough evidence to reject $H_0$ with 95\% confidence. Thus, the T-test states that $H_a$ is indeed true, overlaying that the mean MSE of the proposed method is less than all other methods. This result is also significantly demonstrated in the p-values column, where all values are less than 0.05.

\subsection{Experiments with M4 Competition Datasets}
Here, the hourly M4 dataset is used as a real-life dataset, which
includes 414 different time series data  \citep{MAKRIDAKIS2018802}. We used the M4 competition dataset since it includes sequences suitable for time series forecasting apart from being a well-known benchmark. It inherits business forecasting sequences, i.e., finance, and economics, where the non-stationarity is observed at best. Apart from that, business forecasting tasks generally exhibit high dimensionality with the presence of various stock market features. Given that our proposed approach also applies to wind power prediction with weather and stock market features, this dataset is a suitable choice for our simulations. 

About the structure of the series, the M4 competition dataset does not include date-time indexes. Hence, we give the indexes externally to the dataset to extract
date-related features. The sample size of the train set was reduced to 953, while the test set includes 
48 samples as demonstrated in Table \ref{table:data_info}. Investigating $y_t$ regarding the stationarity test; we found the p-value
to be $0.36856$ on average, which indicates non-stationarity, as demonstrated in Table \ref{table:data_info}.
We give the $y_t$-related features to $\boldsymbol{s}_t^{(1)}$ to be consistent with the synthetic dataset. We transformed the desired data through a min-max scaler. In this setting, we took the $2^{nd}$,  $4^{th}$, $6^{th}$, and $8^{th}$ lags of the desired data based on autocorrelation values to generate the mean and standard deviation of the rolling window feature. The date time features are included in $\boldsymbol{s}_t^{(2)}$. The date-related features are the cosine and sinus vectors of the hour, day of the month, day of the week, month of the year, quarter of the year, and week of the year. Thus, \textbf{the high dimensionality and non-stationarity of the data is validated by the number of features being larger than 30 and p-value being larger than 0.05 after the ADF test.}

\begin{figure}[htbp]
    \centering
    \includegraphics[width=\columnwidth]{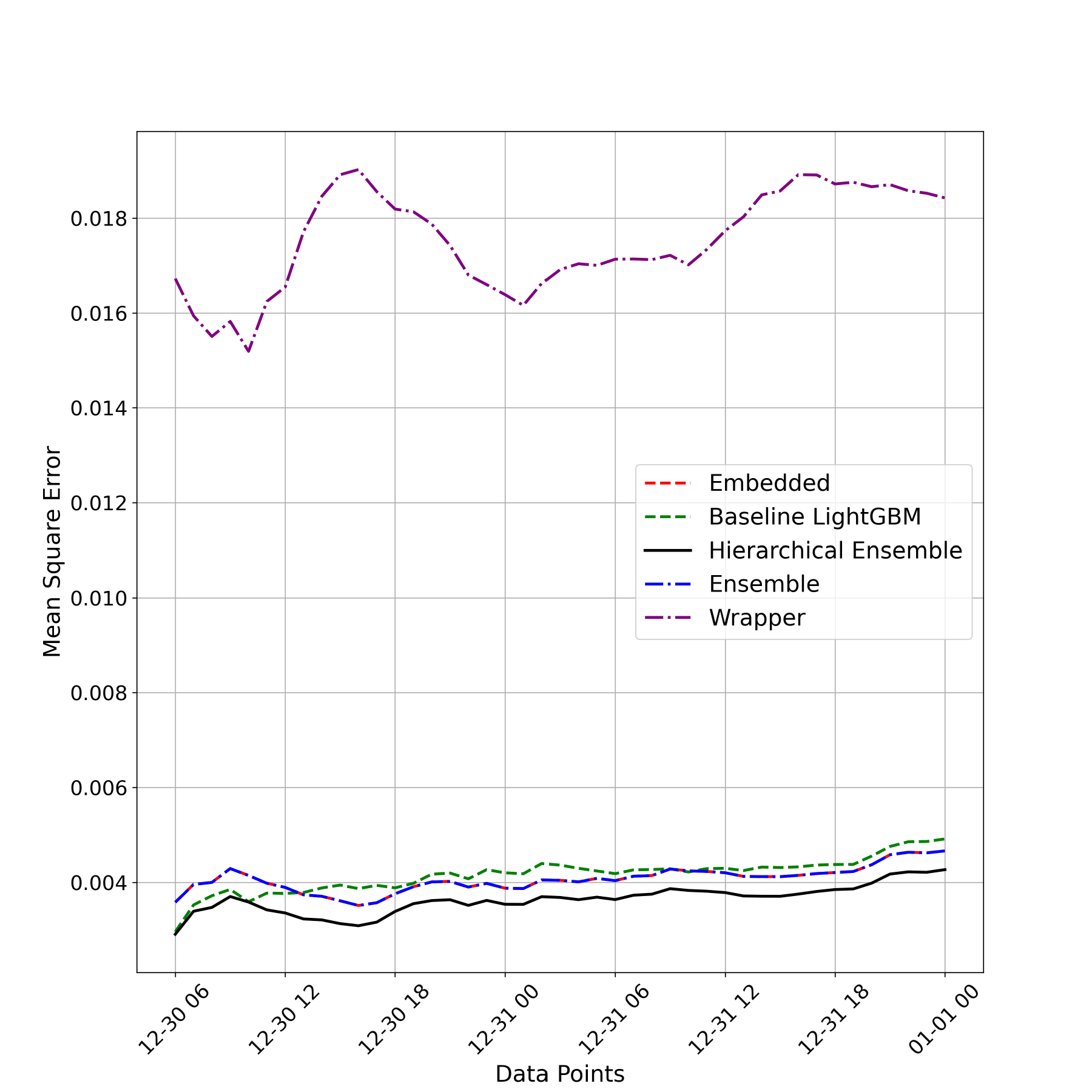}
    \caption{Comparison of the mean square error performances of \emph{Hierarchical Ensemble} (black), \emph{Ensemble} (blue), \emph{Base LightGBM} (green), \emph{Embedded} (red), \emph{Wrapper} (purple) for the M4 hourly competition dataset.}
    \label{fig:m4_results}
\end{figure}

Based on the performance of the proposed method in the M4 hourly dataset in Figure \ref{fig:m4_results}, the
\emph{Hierarchical Ensemble} outperforms other models. We highlight that the $y_t$-related features given in a higher level of hierarchy form the long-term patterns while the feature subset containing domain knowledge easily upscales or downscales the long-term patterns. Moreover, we demonstrate a more robust performance by the stationary cumulative error compared to other methods which are either oscillating. 
The reason that \textit{Baseline LightGBM} performs worse than our method is because of the high dimensionality causing
overfitting. Although the feature size is less than the synthetic experiment, the \textit{Baseline LightGBM} memorizes the unique patterns in the train set due to the highly informative features.

\begin{table}[htbp]
    \caption{The Comparison of Average Time Consumption in Seconds}
    \label{table:data_duration}
        \begin{tabular}{cccc}
            \hline
            \textbf{Dataset} & \textit{Wrapper} & \textit{Hierarchical Ensemble} \\
                        \hline

            Synthetic & 393.8 & 9.3552  \\

            M4 Hourly Forecasting & 75.24 &  17.46 \\
            \hline
        \end{tabular}
\end{table}

The \textit{Embedded}  and \textit{Ensemble}  methods perform close to each other since the \textit{Ensemble} model generally chooses the \textit{Embedded} model giving $\alpha_t$ of 1 among the range of $[0,1]$. It is important to note that the \textit{Embedded} utilizes a more dominant feature set compared to date-related features, which indicates the inclination of \textit{Baseline LightGBM} to $y_t$-related features. Our proposed method overcomes this issue by finetuning with the less dominant features in the last level for a greater impact on $\tilde{y}_t^{(2)}$. 
Moreover, The \textit{Wrapper} method clearly converges to a non-optimal local minimum since it performs the worst. It also demonstrates the least robust MSE performance due to oscillatory behavior in contrast to our method.  From Table \ref{table:data_duration}, the extreme time consumption of this method compared to \textit{Hierarchical Ensemble} also verifies the efficiency of our model. 
As our proposed method performs the best, we conclude that the date-related features modify the first layer predictions successfully, solving overshooting or undershooting problems. Overall, the proposed method works satisfactorily in the real-life dataset outperforming other models.

\begin{table}[htbp]
    \centering
    \caption{Paired t-test results for Mean Squared Error (MSE) - Real Life Data}
    \begin{tabular}{lcc} 
        \toprule
        Method & T-statistic & P-value \\
        \midrule
        Baseline LightGBM & 9.589 & $6.09 \times 10^{-13}$ \\
        Ensemble & 25.171 & $3.15 \times 10^{-29}$ \\
        Embedded & 25.103 & $3.55 \times 10^{-29}$ \\
        Wrapper & 64.487 & $7.93 \times 10^{-48}$ \\
        \bottomrule
    \end{tabular}
    \label{tab:real_data}
\end{table}

To test the statistical significance as introduced in Section \ref{sec:hyperparameter}, we input the first samples of MSE as the compared methods and the second as the proposed method. Table \ref{tab:real_data} demonstrates that we have enough evidence to reject $H_0$ with 95\% confidence. Thus, the T-test states that $H_a$ is indeed true, overlaying that the mean MSE of the proposed method is less than all other methods. This result is also significantly demonstrated in the p-values column, where all values are less than 0.05.

\section{Conclusions}\label{sec:Conclusion}
In this work, we proposed an ensemble feature selection method based on hierarchical stacking.
On top of the important milestones of traditional stacking methods, our approach leverages a hierarchical structure that fully exploits the co-dependency between features. This hierarchical stacking involves training an initial machine learning model using a subset of the features and then updating the output of the model using another machine learning algorithm that takes the remaining features or a subset of them to
adjust the first layer predictions while minimizing a custom loss. This hierarchical structure provides novelty by allowing for flexible depth in each layer and suitability for any loss function. We demonstrate the effectiveness of our approach on the synthetic and M4 competition datasets. Overall, the proposed hierarchical ensemble approach for feature selection offers a robust and scalable solution to the challenges posed by datasets with high dimensional feature sets compared to the sample size. Effectively capturing feature co-dependency and showcasing enhanced accuracy and stability in machine learning
models, our method outperforms traditional and state-of-the-art machine learning models. 
The limitation of this method lies in one of its contributions, where we provide a solution for non-practical second derivatives to GBM algorithms. This contribution can also be tested for practical second derivatives to ensure the feature selection capability and prediction quality.
In future work, the proposed method might go in several directions. The model might be extended to $K>2$ cases to test scalability to multicontext datasets. Moreover, the base models of the hierarchical layers might be extended out of GBMs, i.e., Neural Networks, to test the scalability of any machine learning model. 
We also provide the source code of our approach to facilitate further research and replicability of our results.


\bibliography{refs}

\end{document}